\documentclass[runningheads]{llncs}

\usepackage[T1]{fontenc}
\usepackage{graphicx}
\usepackage{xcolor}
\usepackage{tabularx}
\usepackage{multirow}
\usepackage{hyperref}

\begin{document}

\title{Semi-Supervised Image-Based Narrative Extraction: A Case Study with Historical Photographic Records}

\titlerunning{Semi-Supervised Image-Based Narrative Extraction}

\authorrunning{German et al.}
\author{
    Fausto German\inst{1}\orcidID{0009-0005-0954-4578} \and
    Brian Keith\inst{2}\thanks{Corresponding author}\orcidID{0000-0001-5734-8962} \and
    Mauricio Matus\inst{3}\orcidID{0009-0003-9852-5285} \and
    Diego Urrutia\inst{2}\orcidID{0000-0002-0308-2406} \and
    Claudio Meneses\inst{2}\orcidID{0000-0003-1112-4925}
}

\institute{
    Virginia Tech, Departament of Computer Science\\
    Blacksburg VA 24061, USA\\
    \email{fgermanj@vt.edu}
    \and
    Universidad Católica del Norte, Department of Computing and Systems Engineering\\
    Antofagasta, Chile\\
    \email{\{brian.keith,durrutia,cmeneses\}@ucn.cl}
    \and
    Universidad Católica del Norte, School of Journalism\\
    Antofagasta, Chile\\
    \email{mmatus@ucn.cl}
}

\maketitle
\begin{abstract}
This paper presents a semi-supervised approach to extracting narratives from historical photographic records using an adaptation of the narrative maps algorithm. We extend the original unsupervised text-based method to work with image data, leveraging deep learning techniques for visual feature extraction and similarity computation. Our method is applied to the ROGER dataset, a collection of photographs from the 1928 Sacambaya Expedition in Bolivia captured by Robert Gerstmann. We compare our algorithmically extracted visual narratives with expert-curated timelines of varying lengths (5 to 30 images) to evaluate the effectiveness of our approach. In particular, we use the Dynamic Time Warping (DTW) algorithm to match the extracted narratives with the expert-curated baseline. In addition, we asked an expert on the topic to qualitatively evaluate a representative example of the resulting narratives. Our findings show that the narrative maps approach generally outperforms random sampling for longer timelines (10+ images, p < 0.05), with expert evaluation confirming the historical accuracy and coherence of the extracted narratives. This research contributes to the field of computational analysis of visual cultural heritage, offering new tools for historians, archivists, and digital humanities scholars to explore and understand large-scale image collections. The method's ability to generate meaningful narratives from visual data opens up new possibilities for the study and interpretation of historical events through photographic evidence. Source code and experiments available on \href{https://github.com/faustogerman/ROGER-Concept-Narratives}{GitHub}.

\keywords{Computational Narrative Extraction \and Visual Narratives \and Robert Gerstmann \and Historical Photograph Data \and Narrative Maps}
\end{abstract}
\section{Introduction}
In recent years, the digitization of historical archives has opened up new possibilities for analyzing and understanding our past through visual records \cite{wevers20222}. Photographs, in particular, offer rich insights into historical events, cultural practices, and societal changes. However, the sheer volume of digitized images presents challenges in extracting meaningful narratives from these vast collections \cite{mannisto2022automatic}.

While significant progress has been made in text-based narrative extraction \cite{keith2021narrative,keith2023survey}, the field of image-based narrative construction remains relatively unexplored. This gap is particularly notable in the context of historical photographic records, where the extraction of coherent narratives could provide valuable and efficient insights for historians, archivists, and researchers across various disciplines \cite{arnold2019distant}.

In this context, we conceptualize narratives in this work as both an intentional and retrospective constructs in historical photography collections. Drawing from Ricoeur's concept of narrative temporality \cite{ricoeur1984time} and Edwards' understanding of photographic collections as ``material performances of history'' \cite{edwards2009photography}, we approach the historical photographic records as sequences of images that form coherent stories. 

As a first step toward the goal of algorithmic concept narrative extraction, this paper presents a proof-of-concept application for the extraction of image-based storylines from historical photographic records using a semi-supervised approach with the narrative maps algorithm. In particular, our primary contributions are:

\begin{enumerate}
    \item A novel semi-supervised adaptation of the narrative maps algorithm \cite{keith2021narrative} that leverages domain-specific knowledge through a dataset of images partially labeled by experts.
    \item A demonstration of the feasibility of extracting coherent narratives from a collection of historical photographs from the ROGER dataset \cite{matus2024roger} through qualitative and quantitative evaluations.
\end{enumerate}

This work extends the existing unsupervised text-based approaches to the visual domain in a semi-supervised setting. We present an adaptation of the \textit{Narrative Maps} algorithm \cite{keith2021narrative}, tailoring it to work with visual features and historical photographs with partial labels provided by an expert annotator. Through a case study that applies our method to the ROGER dataset \cite{matus2024roger}, we demonstrate its potential for uncovering hidden narratives in historical photographic collections.

Our research provides valuable insights into the challenges and opportunities presented by image-based narrative extraction in the context of historical research and digital humanities. By bridging the gap between text-based narrative extraction techniques and image-based historical records, this work offers new tools for researchers to uncover and analyze visual narratives in large-scale photographic archives. Our proof-of-concept application showcases the potential of this approach to enhance our understanding of historical events and cultural phenomena through the lens of photographic evidence.

The remainder of this paper is organized as follows: Section 2 reviews related work on the extraction of narratives from text and images, and the analysis of historical photographic records. Section 3 presents our methodology, including the adaptation of the Narrative Maps algorithm for visual data, our semi-supervised labeling approach, and evaluation methods. Section 4 presents both quantitative and qualitative results from applying our method to the ROGER dataset. Section 5 discusses the implications of our findings, limitations, and potential future work. Finally, Section 6 concludes with a summary of our contributions and broader implications for computational analysis of visual cultural heritage.

\section{Related Work}
\subsection{Narrative extraction from text}
Narrative extraction from textual data has been a significant focus in computational linguistics and natural language processing \cite{santana2023survey}. Recent literature reviews provide valuable context around narrative extraction approaches. Although most reviews have focused on narrative generation techniques \cite{alhussain2021automatic}, some work has examined specific narrative extraction subtasks such as fictional character network analysis \cite{labatut2019extraction} and timeline summarization \cite{ghalandari2020examining}. However, in this work, we focus on an \textit{event-based} approach to narrative extraction \cite{keith2023survey}, associating the images from our data to specific events of the narrative.

In general, researchers have developed various event-based approaches to identify and extract coherent storylines from large text corpora \cite{keith2023survey}. In particular, Keith and Mitra \cite{keith2021narrative} introduced the narrative maps algorithm, which uses a graph-based approach to represent and extract information narratives from news articles. Their work builds on earlier efforts, such as the Connect-the-Dots algorithm \cite{shahaf2010connecting}, which aimed to create coherent news storylines. 

The aforementioned method is an \textit{unsupervised} approach that relies on projecting high-dimensional embeddings using dimensionality reduction techniques and clustering to identify the `topics' in the data to assess topical similarity between the narrative events. In this work, we propose a \textit{semi-supervised} approach that leverages expert-provided labels of the images to replace the unsupervised clustering step with a semi-supervised approach.

\subsection{Image-based narrative construction}
While text-based narrative extraction has seen significant advancements, image-based narrative extraction remains a relatively nascent field. However, recent years have seen growing interest in this area. Huang et al. \cite{huang2016visual} proposed a method for visual storytelling, generating textual descriptions for sequences of images. Building on this, Wang et al. \cite{wang2018no} developed a hierarchical attention-based model for visual story generation, incorporating both local and global visual features.

In the context of historical images, Wevers and Smits \cite{wevers2020visual} demonstrated the potential of using neural networks to analyze large collections of historical images, paving the way for more advanced narrative extraction techniques. Arnold and Tilton \cite{arnold2019distant} introduced the concept of ``distant viewing'' for analyzing large visual corpora, emphasizing the need for computational methods in visual cultural heritage studies.

\subsection{Historical photographic record analysis}
The analysis of historical photographic records has gained momentum with the increasing digitization of archives. Männistö et al. \cite{mannisto2022automatic} proposed a framework for automatic image content extraction in large visual archives, addressing the challenges of applying machine learning to humanistic photographic studies. Their work highlights the potential for computational methods to enhance traditional historical research practices.

Chumachenko et al. \cite{chumachenko2020machine} applied deep learning techniques to analyze the work of Finnish World War II photographers, demonstrating the potential of AI in uncovering patterns and narratives in historical photographic collections. The intersection of computer vision and digital humanities has also led to innovative approaches in analyzing historical photographs. Wevers and Smits \cite{wevers2020visual} used transfer learning techniques to classify and analyze advertisements in historical newspapers, showcasing the potential of adapting pre-trained models to historical visual data.

While these studies have made significant strides in applying computational methods to historical visual data, there remains a gap in specifically extracting coherent narratives from collections of historical photographs. Our work aims to bridge this gap by adapting narrative extraction techniques to the visual domain, with a specific focus on historical photographic records.

\section{Methodology}
\begin{figure*}[!htb]
    \includegraphics[width=\textwidth]{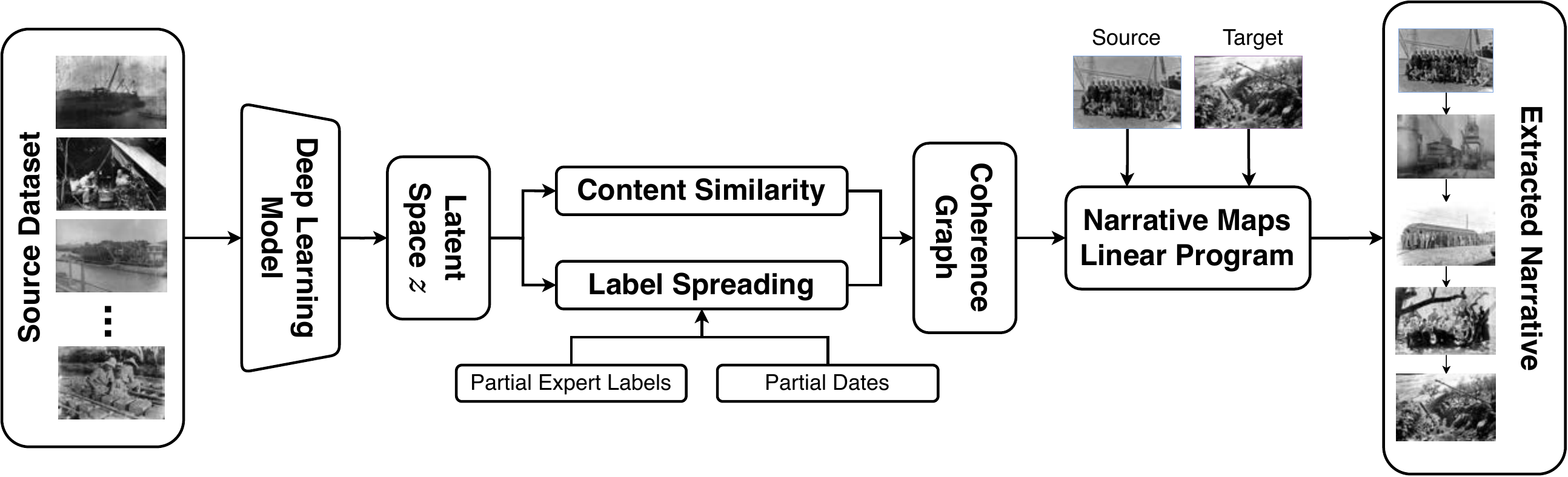}
    \caption{The proposed visual narrative extraction pipeline. We construct a coherence graph based on the content semantic similarity and partial label information of a collection of images. During extraction, users can select source and target images to extract concept narratives using the adapted narrative maps algorithm.}
    \label{fig:extraction-pipeline}
\end{figure*}

Our approach adapts the original unsupervised narrative maps algorithm \cite{keith2021narrative} to work with visual data, specifically historical photographs, in a semi-supervised manner. The method consists of several key steps: data preparation, feature extraction, semi-supervised labeling, similarity computation, narrative extraction, and evaluation. Figure \ref{fig:extraction-pipeline} illustrates the overall pipeline of our approach.

Our proposed workflow for image-based narrative extraction adapts the Narrative Maps algorithm \cite{keith2021narrative} for visual data. We begin by extracting and normalizing visual features from each photograph using a pre-trained DETR (DEtection TRansformer) model \cite{nicolas2020Detr}. These features are then used to compute pairwise similarities between images, creating a similarity matrix. We incorporate temporal information by utilizing expert-provided chronological ordering to create a directed graph, similar to the original algorithm's approach.

To identify visual themes, we incorporate partial thematic labels provided by experts using label-spreading algorithms on the image feature space. We then implement coverage constraints to ensure representation from different thematic clusters, adhering to the original algorithm's emphasis on diverse topic coverage. The narrative map construction follows the optimization framework described in the original paper \cite{keith2021narrative}, maximizing coherence subject to coverage and structural constraints. From this structure, we extract the main route, representing the primary storyline and the main output of our pipeline.

\subsection{Data Preparation and Feature Extraction}
The ROGER (Robert Gerstmann's Expeditions Repository) dataset \cite{matus2024roger} is a collection of historical photographs taken by Robert Gerstmann (1896-1964), a German photographer and engineer who extensively documented South America in the early 20th century. This dataset has been the subject of previous research \cite{matus2024roger}, which focused on extracting narratives from descriptions of the images generated by Large Language Models (LLMs) \cite{makridakis2023large}. However, this approach was limited to a very small subset of the complete data set. 

The complete Gerstmann archive consists of over 47,000 images, spanning approximately 40 years of photographic capture, including 43,475 negatives and 15,054 positives in various formats \cite{alvarado2009roberto}. For this study, we focus on a subset of the collection, specifically the images related to the Sacambaya Expedition of 1928 \cite{sanders1928jesuit}. This is the same subset of the ROGER collection used in preliminary work on LLM-based narrative extraction from images \cite{matus2024roger}. 

In particular, the selection of these images followed a two-stage filtering process. The primary selection stage identified 545 images from the archive that constituted a discrete historical unit, specifically documentation activities conducted between March and November 1928 \cite{jolly1934}. This selection was then restricted to images captured and stored in a uniform 10 x 15 mm flexible film format, although these specimens lack sequential correlation in their original documentation. The secondary selection stage focused on image quality and legibility, leading to the exclusion of 45 images: 36 due to visual aberrations from light overexposure and 9 due to focus and sharpness issues.

This subset includes 500 usable images capturing the five-month treasure-hunting expedition in Bolivia. Each image is preprocessed and resized to a standard dimension (256x256 pixels). For feature extraction, we employ a pre-trained DETR model to generate high-dimensional feature vectors for each image. These feature vectors capture rich visual and conceptual information and serve as the basis for computing similarities between images.

\subsection{Semi-supervised Labeling}
We utilize a semi-supervised approach to propagate category and date labels across the dataset. This process leverages partial expert-provided labels and the extracted visual features. The category labels propagation replaces the original clustering step in the unsupervised narrative maps algorithm \cite{keith2021narrative}. More specifically, we use the cluster probability vectors obtained from the category label propagation to compute the topic similarity between photographs.

\paragraph{Thematic clustering.} While all images in our dataset contained an approximate location tag, only a portion of them were labeled thematically by domain experts (e.g., ``marine transport'', ``motorized land transport''). So, to incorporate thematic similarity into our pipeline, we performed semi-supervised clustering to identify thematic groups within the dataset. This is achieved by concatenating the available location tags with the DETR embeddings and applying a label-spreading algorithm \cite{zhou2003learning} to this augmented feature space using the partial thematic labels as a seed. This clustering provides an initial structure for understanding the visual themes present in the expedition narrative.

\paragraph{Temporal information.} Similar to the expert thematic labels, only a portion of the dataset contained expert-based date approximations for the images. Since the Narrative Maps algorithm assumes a temporal ordering of the dataset, we apply a label-spreading algorithm using the provided expert-based dates as seeds to approximate the date labels in the rest of the dataset.

\subsection{Narrative Maps algorithm}
The Narrative Maps algorithm, introduced by Keith and Mitra \cite{keith2021narrative}, is a graph-based approach for representing and extracting information narratives from textual data. This algorithm is the foundation of our narrative extraction method. The key components of the Narrative Maps algorithm are as follows:

\begin{itemize}
\item \textbf{Graph Representation:} The algorithm represents narratives as a directed acyclic graph with a single source (starting event) and a single sink (ending event). Each node in the graph represents an event, while edges represent the connections between events.
\item \textbf{Coherence Maximization:} The algorithm aims to maximize the coherence of the narrative by optimizing the strength of the weakest link in terms of coherence. This is based on the principle that a story is as coherent as its weakest connection \cite{shahaf2010connecting}. In our semi-supervised version, we redefine coherence in terms of visual similarity and thematic consistency. That is, the coherence between two images is computed using a combination of their DETR feature similarity, representing the match between the underlying concepts captured by the image, and the results of the label spreading algorithm. This concept-based coherence formulation represents the main distinction between our proposed methods and the traditional narrative maps algorithm.
\item \textbf{Coverage Constraints:} To ensure that the extracted narrative covers a diverse range of topics, the algorithm incorporates coverage constraints. These constraints ensure that a certain percentage of the extracted topic clusters are represented in the final narrative. 
\item \textbf{Main Route Extraction:} After constructing the basic narrative map structure, the algorithm identifies the main storyline by finding the most coherent path from the starting event to the ending event. This is computationally achieved by finding the maximum likelihood path in the graph.
\item \textbf{User-defined Parameters:} There are two key user-defined parameters. First, the size of the map is regulated by the $K$ parameter, which represents the expected number of events in the main story. Second, the minimum coverage of the map is regulated by the \textsf{mincover} parameter.
\end{itemize}
 
\subsection{Evaluation}
We evaluated our method using expert-curated timelines of varying lengths (5 to 30 images) as the ground truth. For each timeline length, we perform multiple extractions with different random shuffles of the dataset to ensure robustness. In particular, we use two primary metrics for evaluation: \textbf{Dynamic Time Warping} (DTW) Distance \cite{muller2007dynamic}, which measures the similarity between the extracted narrative and the expert-curated timeline by sequentially matching images in both sets while allowing for non-linear distortions along the sequence dimension, and \textbf{Average Cosine Similarity}, which compute the average cosine similarity between matched images along the optimal matching path obtained from the DTW algorithm. We compare our Narrative Maps (NM) approach against a random sampling (RS) baseline to assess the effectiveness of our method.

\subsection{Expert-curated baselines}
\begin{figure}[!htb]
    \centering
    \includegraphics[width=1\linewidth]{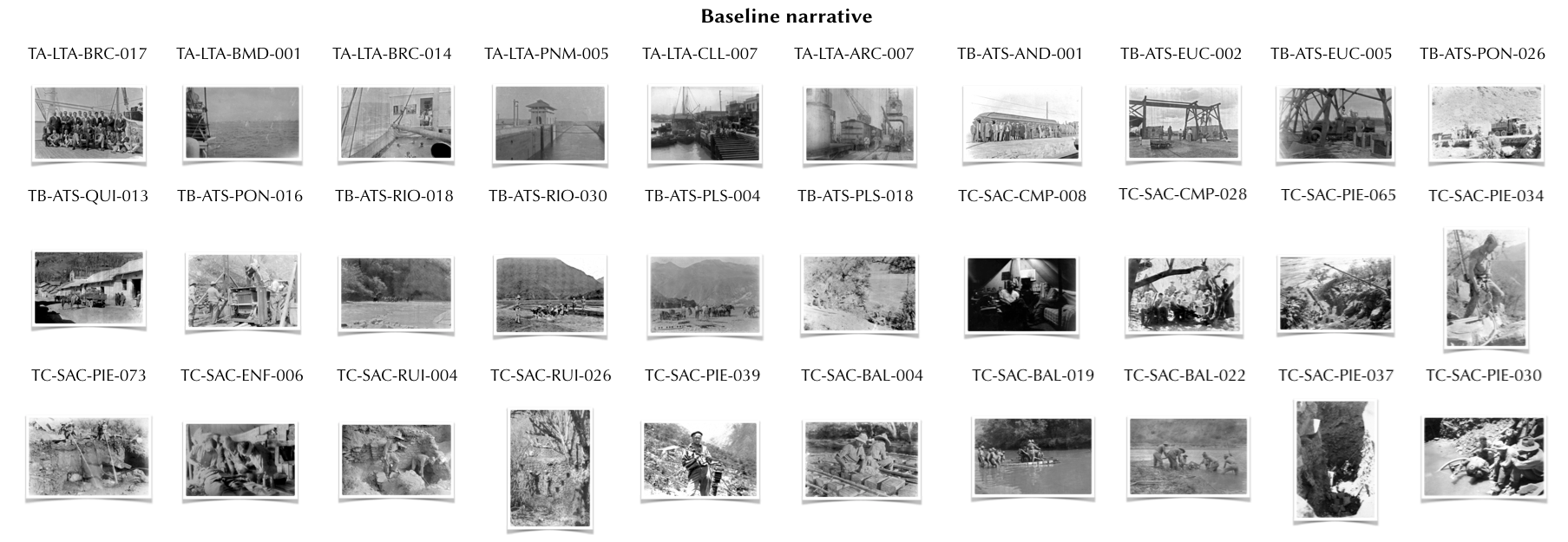}
    \caption{Expert-curated narrative used as the baseline for evaluation, with images arranged in English reading order (left to right, top to bottom).}
    \label{fig:baseline}
\end{figure}

To evaluate our narrative extraction method, we used expert-curated timelines as baselines. A domain expert with deep knowledge of the Sacambaya Expedition created six timelines of varying lengths: 5, 10, 15, 20, 25, and 30 images. These timelines were constructed using historical references and the original documentary corpus of images from the Gerstmann collection. Figure \ref{fig:baseline} illustrates one of the expert-curated baseline with 30 images.

The expert chronologically arranged the images, starting from March 1928 when the research team departed from England. Key milestones of the journey, including various modes of transport and significant locations, were correlated with accounts from primary texts and contemporary newspaper reports. The expert labeled the images according to expedition stages, using categories such as \textit{marine transport}, \textit{motorized land transport}, \textit{main excavation site}, \textit{human-animal land transport}, \textit{alternative excavation site}. 

More specifically, all image files were divided into three overarching categories, labeled A, B, and C to indicate chronological order. Within these overarching categories, subcategories were created to represent both inter-locality transfers (e.g., ``Liverpool to Arica,'' abbreviated as ``LTA'') and specific geographical locations (e.g., the excavation site ``Sacambaya,'' designated as ``SAC''). To further refine the classification, tertiary identifiers were implemented to denote specific sites within these broader locations. For instance, within the ``SAC'' subset of photographs, distinct areas such as the workshop (TLL) and the expedition members' camp (CMP) were delineated. This multi-tiered labeling approach was systematically applied across the entire photographic corpus, resulting in the establishment of 21 distinct thematic clusters encompassing the complete set of 500 photographs.

We note that these materials lacked contextual information and chronological order due to various factors: their original capture, their subsequent archiving by the author (1928-1964) and later interventions by the collection's custodians (1964-present). To construct the baselines, the expert relied on two main textual sources: a primary text of a narrative nature \cite{jolly1934} and a more recent secondary text providing historical systematization, supported by contemporary newspaper accounts of the events \cite{condori2015}. By applying a quantitative linguistic approach \cite{fleischman1990tense} to the primary text with narrative sequencing based on temporal markers, the expert established a basic chronology of the expedition's main events. 

These expert-curated timelines serve as ground truth sequences against which we compare our algorithmically extracted narratives, providing a robust basis for evaluating the performance of our method. Furthermore, the expert labeled the images in the baselines according to the stages of the expedition. These partial labels and dates provided by the expert are fed to the semi-supervised narrative maps extraction pipeline.

\subsection{Experimental setup}
Our experimental design aims to evaluate the performance of our semi-supervised Narrative Maps (NM) approach against a Random Sampling (RS) baseline across various narrative lengths. We use the expert-curated timelines of 5, 10, 15, 20, 25, and 30 images as ground truth for comparison.

For each timeline length, we conduct 20 independent trials. In each trial, we randomly shuffle the dataset while keeping the start and end images fixed to match the expert-curated timeline. We then extract narratives using both our NM approach and the RS baseline and evaluate the extracted narratives using the DTW and Cosine Similarity metrics.

To assess the impact of dimensionality, we perform this evaluation in both high-dimensional (original DETR embeddings) and low-dimensional (UMAP-reduced) spaces. We used UMAP \cite{mcinnes2018umap} as this method was part of the original unsupervised Narrative Maps extraction pipeline \cite{keith2021narrative}. This allows us to compare the performance of our method in different feature representations. 

For statistical analysis, we use t-tests to compare the performance of NM against RS for each timeline length and each metric. We consider results statistically significant at p < 0.05. This approach allows us to determine whether our NM method consistently outperforms random selection across different narrative lengths and feature spaces.

We note that all baselines use the same starting and ending images, except the timeline of length 5. Thus, in most cases, we use the same images as fixed points in our extraction process. To regulate the extraction process and enable comparison with baselines of different lengths, we utilize the parameter $K$ of the narrative maps extraction algorithm, which represents the expected length of the main story \cite{keith2021narrative,shahaf2010connecting}. We vary $K$ to match the lengths of our baseline timelines, including the fixed starting and ending points. These restrictions ensure a fair comparison between our method and the expert-curated timelines.

Finally, we complement our quantitative evaluation with an expert-based qualitative evaluation. To do this, we engage with the domain expert who created the baseline timelines to evaluate our extracted narratives. The expert assesses the coherence, relevance, and historical accuracy of one of our algorithmically generated storylines compared to their manually curated timelines.

\section{Results and Discussion}
\subsection{Quantitative evaluation: Similarity to the Baselines}
We now present the results of our semi-supervised image-based narrative extraction method using the ROGER dataset. Table \ref{tab:similarity_scores} shows the average similarity scores and Dynamic Time Warping distance for each baseline timeline length and each method (NM and RS). 

\begin{table}[ht!]
\caption{Comparison of Dynamic Time Warping similarity and distance between Narrative Maps and random samples using the expert-generated baselines of different lengths as ground truth, where $L$ is the length of the baseline timelines.}
\label{tab:similarity_scores}

\centering
\scalebox{0.73}{
\begin{tabular}{c|ccc|ccc||ccc|ccc}
\hline
\multirow{2}{*}{$L$} 
 & \multicolumn{3}{c|}{\textbf{High-Dim Similarity}} 
 & \multicolumn{3}{c||}{\textbf{High-Dim Distance}}
 & \multicolumn{3}{c|}{\textbf{Low-Dim Similarity}} 
 & \multicolumn{3}{c}{\textbf{Low-Dim Distance}} 
\\
\cline{2-13}
 & NM & RS & p-value 
 & NM & RS & p-value 
 & NM & RS & p-value
 & NM & RS & p-value
\\
\hline
5  & 0.823 & 0.841 & 0.3125 & 2.208 & 2.791 & 0.0838 & 0.126 & 0.094 & 0.7666 & 8.348 & 8.331 & 0.9821 \\
10 & 0.915 & 0.895 & 0.0399 & 2.313 & 2.615 & 0.0681 & 0.299 & 0.116 & $2.5\mathrm{e}{-4}$ & 8.691 & 10.495 & $8.2\mathrm{e}{-4}$ \\
15 & 0.950 & 0.936 & 0.0257 & 2.215 & 2.424 & 0.0516 & 0.331 & 0.189 & $6.4\mathrm{e}{-4}$ & 8.800 & 10.625 & $6.1\mathrm{e}{-5}$ \\
20 & 0.963 & 0.945 & 0.0033 & 2.108 & 2.401 & 0.0079 & 0.398 & 0.252 & $5\mathrm{e}{-5}$ & 8.339 & 10.086 & $3.4\mathrm{e}{-5}$ \\
25 & 0.960 & 0.954 & 0.3514 & 1.867 & 2.408 & $1.2\mathrm{e}{-4}$ & 0.480 & 0.261 & $1.6\mathrm{e}{-11}$ & 7.258 & 10.031 & $1.9\mathrm{e}{-11}$ \\
30 & 0.969 & 0.956 & 0.0087 & 2.026 & 2.460 & $2.1\mathrm{e}{-5}$ & 0.481 & 0.279 & $1.1\mathrm{e}{-9}$ & 7.327 & 10.029 & $9.4\mathrm{e}{-11}$ \\
\hline
\end{tabular}
}
\end{table}

We note that we do not explicitly evaluate coherence, because the Narrative Maps approach seeks to maximize the minimum coherence \cite{keith2021narrative} and thus the evaluation would be biased in favor of the Narrative Maps. However, we do note that in general, average coherence values of the extracted NM are higher compared to the random sampling with respect to the expert-curated baseline. These differences in coherence are statistically significant except in the smallest baseline of length 5, where there is no significant difference.

\textbf{High-dimensional Embedding Analysis}: The DTW distance results show that our NM approach generally outperformed the RS baseline, particularly for longer timelines. In general, the NM approach consistently achieved lower DTW distances, indicating better alignment with expert-curated timelines. The improvement was statistically significant for timelines of 20+ photos. In terms of average cosine similarity, our NM approach showed better performance, except for the timelines of length 5 and 25. However, adjusting for multiple comparisons, the results in terms of mean similarity are generally only marginal improvements over the RS baseline.

\textbf{Low-dimensional Embedding Analysis}: We also evaluated our method using low-dimensional embeddings obtained through UMAP dimensionality reduction. In the low-dimensional space, our NM approach showed significantly better results compared to the RS baseline for timelines of length 10+. The NM approach achieved lower DTW distances and mean similarities. However, for the shortest timeline, the differences were not statistically significant.

\subsection{Qualitative evaluation: Expert-based Evaluation}
We now present the results of our qualitative evaluation. We show an example narrative extracted by our method in Figure \ref{fig:extracted-narrative}. The feedback of the domain expert on our extracted visual narratives provides valuable insights. In particular:

\begin{figure}[!htb]
    \centering
    \includegraphics[width=\linewidth]{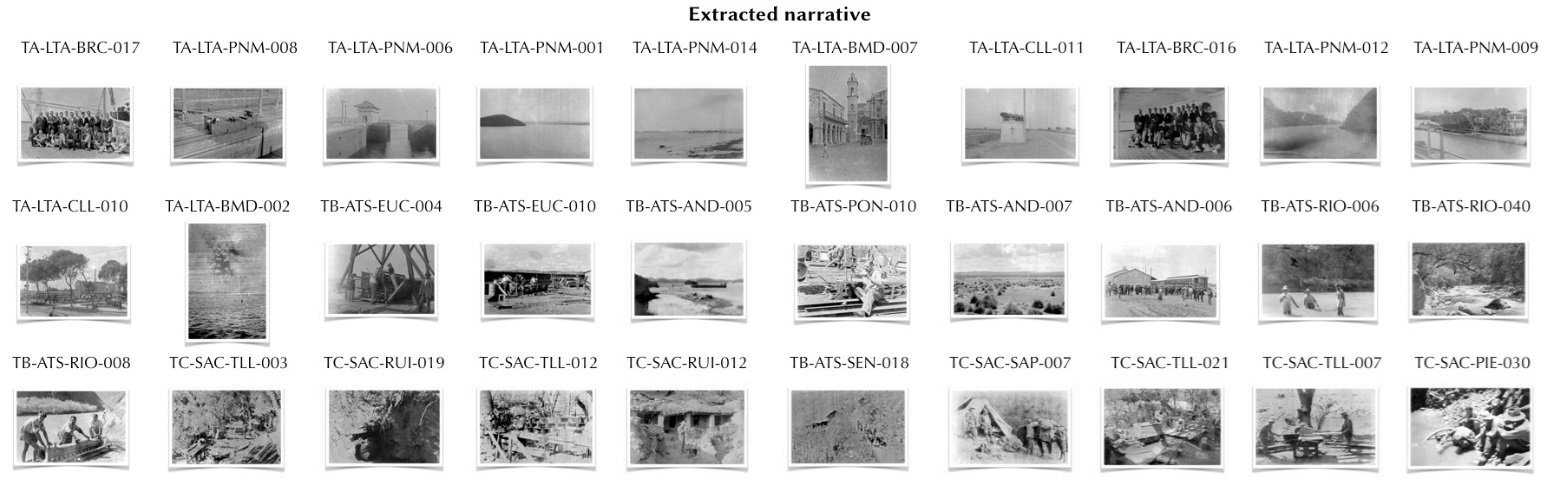}
    \caption{Example extracted narrative using the unsupervised narrative maps algorithm, with images arranged in English reading order (left to right, top to bottom).}
    \label{fig:extracted-narrative}
\end{figure}

\textbf{Coherence.} While there is a lack of exact one-to-one correspondence between the images selected by the human expert and those chosen by the algorithm --- with the intentional exception of the entry and exit points --- an implicit coherence is discernible in the chronological structure of the extracted narrative timeline. This underlying coherence is predicated on the correlations between the aforementioned clusters and, more significantly, on the sequential presentation of the cluster contents, which aligns with the established baseline chronology.

\textbf{Relevance.} The relevance of the algorithmically selected images demonstrates congruence with the baseline proposal. Notably, in certain instances, the algorithm's selections could be interpreted as possessing greater relevance from an objective, image-content-based perspective. This observation highlights the potential divergence between computational objectivity and the contextualized interpretations of a human expert immersed in a specific research framework.

\textbf{Historical Accuracy.} The extracted narrative timeline exhibits valuable historical fidelity in its organization of content. This accuracy is manifested through a narrative coherence that corresponds to the factual sequence of events, substantiated by textual records and historical documentation. The alignment between the computational output and historical reality underscores the potential of this approach in digital heritage studies. We note that minor adjustments to the timeline could be implemented to maintain human oversight in the process without significantly impacting the overall results.

\section{Discussion}
Our results demonstrate that the semi-supervised Narrative Maps approach is generally effective in extracting coherent visual narratives from historical photograph collections. In particular, the improved performance on longer timelines (15 images and above) indicates that our method is especially suited for extracting extended narratives, where coherence and structure become more crucial. 

It is worth noting that the random sampling baseline performed surprisingly well in the high dimensional space in terms of similarity and particularly for shorter timelines (length 5). This could be due to the inherent structure and temporal coherence present in the ROGER dataset, where even random selections might capture some meaningful sequences. In particular, given the relatively \textbf{small size} and \textbf{highly focused nature} of the data set, despite representing different stages of the expedition, most images are highly similar. Thus, both NM and RS could provide a decent overview of the data set according to the results provided by the evaluation metrics. In this context, both approaches tend to emulate the underlying class distribution of the data, which is similar to how the narrative baseline covers the different stages of the expedition. These results align with previous work on evaluating the capabilities of the Narrative Maps algorithm to capture the underlying class distributions of a data set regardless of the choice of starting and ending events \cite{concha2024framing}.

To test the effects of the data distribution on the Narrative Maps algorithm, we weighted the coherence values by the inverse topic frequency to prevent the model from following the data distribution too strongly. That is, for each edge $(i, j)$ in the coherence graph, the edge weight is multiplied by the inverse topic frequency of node $j$. However, this did not provide a statistically significant difference to the model. In later versions of our model, we dropped this capability altogether as it prevented the Narrative Maps model from finding meaningful storylines between the predefined endpoint photographs. From these observations, we note that in \textbf{noisier data sets}, it is likely that the coherence-based approach would provide a better result when compared against random sampling.

Nevertheless, our semi-supervised approach demonstrates promise in extracting meaningful visual narratives from historical photograph collections, particularly when dealing with longer, more complex storylines. The method's ability to leverage both expert knowledge (through partial labels) and deep learning-based feature extraction contributes to its effectiveness in this domain.

\subsection{Limitations and future work}
Our approach, while promising, faces certain limitations. The reliance on pre-trained image embeddings may not fully capture the nuances of historical photographs, potentially missing specific visual elements from this domain. Future work could address these limitations by exploring domain-specific fine-tuning of visual feature extractors. 

We also note that, while the qualitative nature of our evaluation is appropriate for assessing historical narrative coherence, future work could complement this with additional quantitative metrics. Moreover, while evaluating our methods against random sampling provides us with limited feedback, other potential path-finding baselines fail at this task since the underlying graph structure of our dataset and the extraction objective prevent them from finding storylines beyond the direct connection from the source to the target photographs.

One limitation of the original narrative maps method is that the computational complexity of its linear program presents challenges for scaling to very large data collections \cite{keith2021narrative,keith2023survey}. Although our evaluations on a relatively small dataset of 500 photographs take approximately 15 minutes on a mid-range laptop with 16 GB of unified memory for all experiments, developing more efficient extraction models for image-based narratives on larger datasets presents one possible avenue for future research.

Integrating multimodal data \cite{derby2018using}, such as combining visual features with textual metadata or captions, could further enhance the narrative extraction process. Exploring the application of this method to other types of visual narratives, beyond historical photographs, also presents another avenue for future research.

Moreover, while reliance on expert-based labels is key to our semi-supervised approach's effectiveness, it does limit generalizability to collections where such expertise is unavailable. Future work could explore methods to adapt the framework for varying levels of available expert knowledge.

Furthermore, the current method's dependency on expert-provided labels to induce a temporal ordering could introduce biases or inaccuracies in the extracted narratives. Mitigating and controlling the effects of bias introduced by users in these AI-based narratives remains an open problem \cite{keith2023survey}.

Finally, we note that our method works on the assumption that the data set contains an underlying \textit{narrative}. However, we cannot definitively establish Gerstmann's narrative intentions for the Sacambaya collection. In this context, we argue that the narrative structure identified in this collection represents both an intentional documentary strategy by Gerstmann and what Schwartz et al. \cite{schwartz2002archives} terms a ``historical performance'' --- a deliberate act of heritage documentation that acquires additional narrative layers through subsequent historical analysis.

\section{Conclusions}
This paper presents a semi-supervised adaptation of the narrative maps algorithm for extracting meaningful visual narratives from historical photographic collections. Our approach, which leverages deep learning techniques for feature extraction and incorporates expert knowledge through partially labeled data, demonstrates potential in uncovering coherent visual storylines from the ROGER dataset of the 1928 Sacambaya Expedition.

Experiments show that our Narrative Maps approach outperforms random sampling, particularly for longer timelines and when using high-dimensional DETR embeddings. The method effectively utilizes semi-supervised label spreading for category and date estimation, addressing the challenge of partially labeled historical datasets. However, mixed results with low-dimensional embeddings suggest a trade-off between computational efficiency and narrative quality.

While promising, our work reveals limitations that point to future research directions. These include improving dimensionality reduction techniques, enhancing scalability for larger datasets, and exploring multimodal data integration. As digital archives continue to expand, tools like ours will become increasingly valuable for making sense of visual cultural heritage.

Our semi-supervised adaptation of narrative maps opens up new possibilities for computational narrative extraction in historical photography. By combining deep learning, expert knowledge, and graph-based narrative construction, our method provides a powerful tool for uncovering and analyzing visual stories hidden within large photographic archives, contributing to the broader goal of leveraging artificial intelligence in historical research and cultural preservation.

\section*{Acknowledgments}
This article was supported by Project 202311010033-VRIDT-UCN and Núcleos-202412010049-VRIDT-UCN. We also thank the \textit{Robert Gerstmann Fonds} at Universidad Católica del Norte for providing access to the photographic archive.

%
%
%
\bibliographystyle{splncs04}
\bibliography{bibliography}

\begin{thebibliography}{10}
\providecommand{\url}[1]{\texttt{#1}}
\providecommand{\urlprefix}{URL }
\providecommand{\doi}[1]{https://doi.org/#1}

\bibitem{alhussain2021automatic}
Alhussain, A.I., Azmi, A.M.: Automatic story generation: A survey of approaches. ACM Computing Surveys (CSUR)  \textbf{54}(5),  1--38 (2021)

\bibitem{alvarado2009roberto}
Alvarado, M., Matthews, M., M{\"o}ller, C., Gerstmann, R.: {Roberto Gerstmann}: Fotograf{\'\i}as, paisajes y territorios latinoamericanos. Pehu{\'e}n (2009)

\bibitem{arnold2019distant}
Arnold, T., Tilton, L.: Distant viewing: analyzing large visual corpora. Digital Scholarship in the Humanities  \textbf{34}(Supplement\_1),  i3--i16 (2019)

\bibitem{nicolas2020Detr}
Carion, N., Massa, F., Synnaeve, G., Usunier, N., Kirillov, A., Zagoruyko, S.: End-to-end object detection with transformers. In: Vedaldi, A., Bischof, H., Brox, T., Frahm, J.M. (eds.) Computer Vision -- ECCV 2020. pp. 213--229. Springer International Publishing, Cham (2020)

\bibitem{chumachenko2020machine}
Chumachenko, K., M{\"a}nnist{\"o}, A., Iosifidis, A., Raitoharju, J.: Machine learning based analysis of finnish world war ii photographers. IEEE Access  \textbf{8},  144184--144196 (2020)

\bibitem{concha2024framing}
Concha~Mac{\'\i}as, S., Keith~Norambuena, B.: Evaluating the ability of computationally extracted narrative maps to encode media framing. In: Text2Story@ ECIR. pp. 17--28 (2024)

\bibitem{derby2018using}
Derby, S., Miller, P., Murphy, B., Devereux, B.: Using sparse semantic embeddings learned from multimodal text and image data to model human conceptual knowledge. In: Proceedings of the 22nd Conference on Computational Natural Language Learning. pp. 260--270 (2018)

\bibitem{edwards2009photography}
Edwards, E.: Photography and the material performance of the past. History and Theory  \textbf{48}(4),  130--150 (2009)

\bibitem{fleischman1990tense}
Fleischman, S.: Tense and narrativity: From medieval performance to modern fiction. University of Texas Press (1990)

\bibitem{ghalandari2020examining}
Ghalandari, D.G., Ifrim, G.: Examining the state-of-the-art in news timeline summarization. arXiv preprint arXiv:2005.10107  (2020)

\bibitem{huang2016visual}
Huang, T.H., Ferraro, F., Mostafazadeh, N., Misra, I., Agrawal, A., Devlin, J., Girshick, R., He, X., Kohli, P., Batra, D., et~al.: Visual storytelling. In: Proceedings of the 2016 conference of the North American chapter of the association for computational linguistics: Human language technologies. pp. 1233--1239 (2016)

\bibitem{jolly1934}
Jolly, S.D.: The Treasure Trail. John Long, London (1934)

\bibitem{keith2021narrative}
Keith~Norambuena, B.F., Mitra, T.: Narrative maps: An algorithmic approach to represent and extract information narratives. Proceedings of the ACM on Human-Computer Interaction  \textbf{4}(CSCW3),  1--33 (2021)

\bibitem{keith2023survey}
Keith~Norambuena, B.F., Mitra, T., North, C.: A survey on event-based news narrative extraction. ACM Computing Surveys  \textbf{55}(14s),  1--39 (2023)

\bibitem{labatut2019extraction}
Labatut, V., Bost, X.: Extraction and analysis of fictional character networks: A survey. ACM Computing Surveys (CSUR)  \textbf{52}(5),  1--40 (2019)

\bibitem{makridakis2023large}
Makridakis, S., Petropoulos, F., Kang, Y.: Large language models: Their success and impact. Forecasting  \textbf{5}(3),  536--549 (2023)

\bibitem{mannisto2022automatic}
M{\"a}nnist{\"o}, A., Seker, M., Iosifidis, A., Raitoharju, J.: Automatic image content extraction: Operationalizing machine learning in humanistic photographic studies of large visual archives. arXiv preprint arXiv:2204.02149  (2022)

\bibitem{matus2024roger}
Matus, M., Urrutia, D., Meneses, C., Keith, B.: {ROGER}: Extracting narratives using large language models from {Robert Gerstmann}'s historical photo archive of the {Sacambaya} expedition in 1928. In: Text2Story@ ECIR. pp. 53--64 (2024)

\bibitem{mcinnes2018umap}
McInnes, L., Healy, J., Melville, J.: Umap: Uniform manifold approximation and projection for dimension reduction. arXiv preprint arXiv:1802.03426  (2018)

\bibitem{muller2007dynamic}
M{\"u}ller, M.: Dynamic time warping. Information retrieval for music and motion pp. 69--84 (2007)

\bibitem{condori2015}
Quisbert~Condori, P.: Entre ingenieros y aventureros. {Robert Gerstmann} y el tesoro de {Sacambaya}. In: Imágenes de la Revolución Industrial: {Robert Gerstmann} en las Minas de Bolivia (1925 - 1936), chap.~3, pp. 47--64. Plural Editores, La Paz, Bolivia (2015)

\bibitem{ricoeur1984time}
Ricoeur, P., Ricoeur, P.: Time and narrative, vol.~3. University of Chicago press (1984)

\bibitem{sanders1928jesuit}
Sanders, E.: The Story of the Jesuit Gold Mines in Bolivia and of the Treasure Hidden by the Sacambaya River. Rauner Special Collections Library, Darthmouth College (1928)

\bibitem{santana2023survey}
Santana, B., Campos, R., Amorim, E., Jorge, A., Silvano, P., Nunes, S.: A survey on narrative extraction from textual data. Artificial Intelligence Review  \textbf{56}(8),  8393--8435 (2023)

\bibitem{schwartz2002archives}
Schwartz, J.M., Cook, T.: Archives, records, and power: The making of modern memory. Archival science  \textbf{2},  1--19 (2002)

\bibitem{shahaf2010connecting}
Shahaf, D., Guestrin, C.: Connecting the dots between news articles. In: Proceedings of the 16th ACM SIGKDD international conference on Knowledge discovery and data mining. pp. 623--632 (2010)

\bibitem{wang2018no}
Wang, X., Chen, W., Wang, Y.F., Wang, W.Y.: No metrics are perfect: Adversarial reward learning for visual storytelling. In: Proceedings of the 56th Annual Meeting of the Association for Computational Linguistics (Volume 1: Long Papers). pp. 899--909 (2018)

\bibitem{wevers2020visual}
Wevers, M., Smits, T.: The visual digital turn: Using neural networks to study historical images. Digital Scholarship in the Humanities  \textbf{35}(1),  194--207 (2020)

\bibitem{wevers20222}
Wevers, M., Vriend, N., De~Bruin, A.: What to do with 2.000. 000 historical press photos? the challenges and opportunities of applying a scene detection algorithm to a digitised press photo collection. TMG Journal for Media History  \textbf{25}(1), ~1 (2022)

\bibitem{zhou2003learning}
Zhou, D., Bousquet, O., Lal, T., Weston, J., Sch{\"o}lkopf, B.: Learning with local and global consistency. Advances in neural information processing systems  \textbf{16} (2003)

\end{thebibliography}
\end{document}